\documentclass[letterpaper, 10 pt, conference]{ieeeconf}  

\IEEEoverridecommandlockouts                              

\IEEEoverridecommandlockouts

\usepackage{tikz}
\usepackage{textcomp}
\usepackage{hyperref}
\usepackage{lipsum}

\usepackage{graphicx}
\usepackage{color}

\usepackage{amssymb}
\usepackage{placeins}
\usepackage{amssymb, amsfonts, fouriernc}
\usepackage{amsmath}

\usepackage{dblfloatfix}    

\usepackage{amsmath}
\usepackage{amssymb}
\usepackage[noadjust]{cite}
\newcommand{\R}{\mathbb{R}}
\newcommand{\N}{\mathbb{N}}

\usepackage{fancyhdr,graphicx,amsmath,amssymb}

\usepackage[ruled,vlined]{algorithm2e}
\usepackage{textgreek}
\usepackage{tcolorbox}
\usepackage{amssymb}

\usepackage[noadjust]{cite}

\makeatletter
\newcommand{\removelatexerror}{\let\@latex@error\@gobble}
\makeatother

\renewcommand{\vector}[1]{[x,y,z,q_{w},q_{x},q_{y},q_{z}]^T }
\newcommand{\hatvector}[1]{[\hat{x},\hat{y},\hat{z},\hat{q}_{w},\hat{q}_{x},\hat{q}_{y},\hat{q}_{z}]^T } 
\newcommand{\imuvector}[1]{[\tau_t,a_{x},a_{y},a_{z}, \omega_{x}, \omega_{y}, \omega_{z}]^T}
\newcommand{\desvec}[1]{[x_{d},y_{d},z_{d}]^T}
\newcommand{\desvecvel}[1]{[u_{d},v_{d},w_{d}]^T}

\newcommand{\xdespos}[1]{\mathit{x}_{d}^{p}}
\newcommand{\xdesvel}[1]{\mathit{x}_{d}^{v}}
\newcommand{\dotxdespos}[1]{\mathit{\dot{x}}_{d}^{p}}

\newcommand{\dotxdesvel}[1]{\mathit{\dot{x}}_{d}^{v}}
\newcommand{\ucom}[1]{u_{com}}
\newcommand{\hatxt}[1]{\mathit{\hat{x}}_{t}}
\newcommand{\xt}[1]{\mathit{x}_{t}}
\newcommand{\xprev}[1]{\mathit{x}_{t-1}}
\newcommand{\zt}[1]{z_{t}}
\newcommand{\hprev}[1]{h_{t-1}}
\newcommand{\vomega}[1]{(v_t,\omega_t)}
\newcommand{\yt}[1]{y_t}

\newcommand{\Iimu}[1]{y_{I}}
\newcommand{\Iv}[1]{y_V}

 \newcommand{\qvec}[1]{[q_w, q_x,q_y,q_z]^T}
  \newcommand{\xyzvec}[1]{[x, y,z]^T}

 \newcommand{\qb}[1]{q_t}
\newcommand{\ucomvec}[1]{[F_d,\phi_d, \theta_d, \psi_d]^T}

\title{\LARGE \bf
Learning Pose Estimation for UAV Autonomous Navigation and Landing Using Visual-Inertial Sensor Data}

\author{Francesca Baldini$^{1}$, Animashree Anandkumar$^{1}$, and Richard M. Murray$^{1}$
\thanks{F.Baldini is supported in part by Darpa PAI grant HR0011-18-9-0035. A. Anandkumar is supported in part by Darpa PAI grant HR0011-18-9-0035, Bren Endowed Chair, Microsoft Faculty Fellowship, Google Faculty Award, Adobe Grant, }
\thanks{$^{1}$California Institute of Technology, Pasadena CA 91125, USA}%
\thanks{\textbf{Github}: $\mathtt{https://github.com/Baldins/Learning_UAV_Pose_Estimation}$}
}

\begin{document}

\maketitle
\thispagestyle{empty}
\pagestyle{empty}

\begin{abstract}
In this work, we propose a new learning approach for autonomous navigation and landing of an Unmanned-Aerial-Vehicle (UAV). We develop a multimodal fusion of deep neural architectures for visual-inertial odometry. We train the model in an end-to-end fashion to estimate the current vehicle pose from streams of visual and inertial measurements.
We first evaluate the accuracy of our estimation by comparing the prediction of the model to traditional algorithms on the publicly available EuRoC MAV dataset. The results illustrate a $25 \%$ improvement in estimation accuracy over the baseline. Finally, we integrate the architecture in the closed-loop flight control system of Airsim - a plugin simulator for Unreal Engine - and we provide simulation results for autonomous navigation and landing.
\end{abstract}

\section{INTRODUCTION}
\begin{figure*}[h]
\centering
      \framebox{\parbox{6in}{
      \centering
     \includegraphics[width=\linewidth]{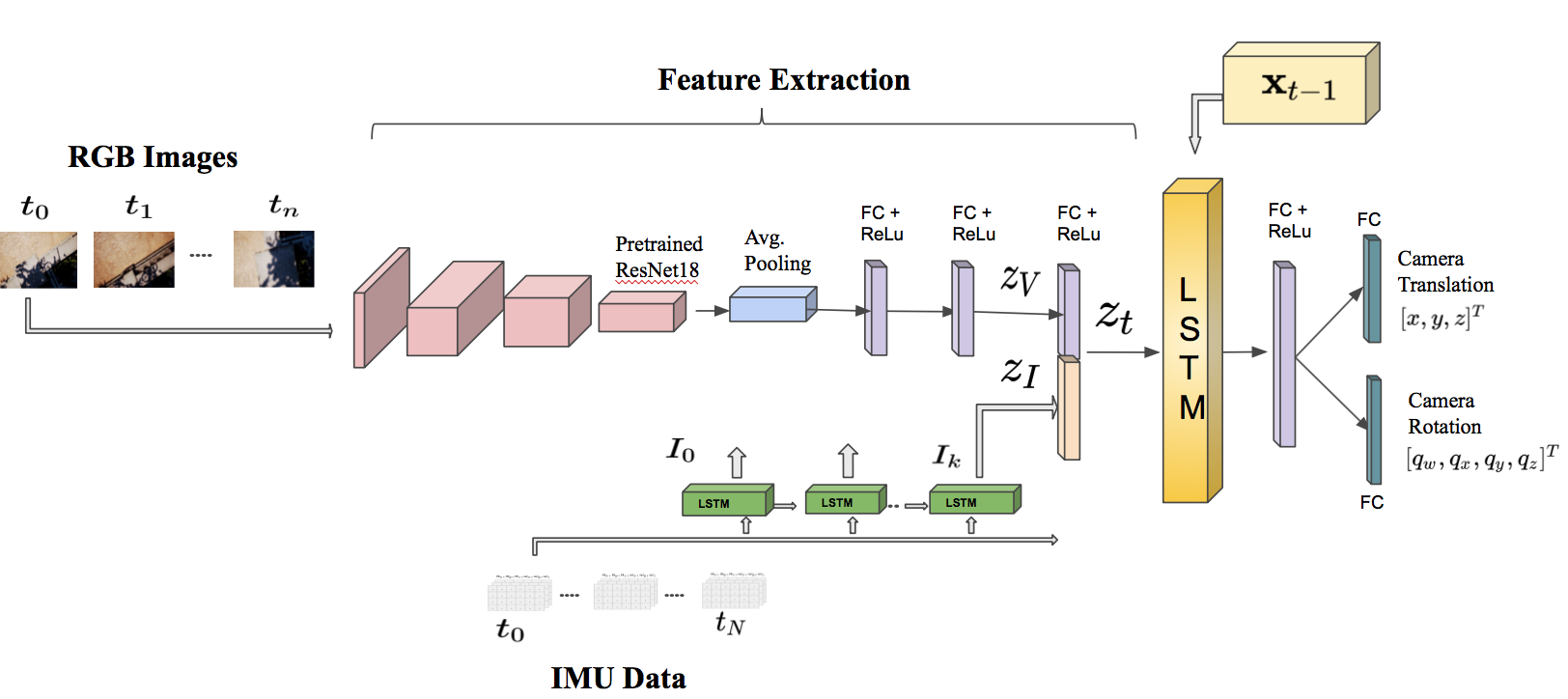}
}}
\caption{Architecture for the data-driven VIO module. It consists of visual and inertial encoders, feature concatenations, temporal modeling, and pose regression. The CNN module determines the most discriminative visual features $z_V$. A small LSTM module transforms windows batch of inertial measurements collected between two consecutive frames into a single inertial feature vector $z_I$. The visual and inertial feature vectors are then concatenated in a single representation $z_t$. The core LSTM uses the feature vector $z_t$ along with the previous estimate $x_{t-1}$ and makes a prediction about the robot pose (translation and rotation).}
      \label{architecture}
\end{figure*}

Unmanned-Aerial-Vehicles (UAVs) can provide significant support for several applications, e.g., rescue operations, environmental monitoring, package delivery, and surveillance, to name a few. 
To guarantee a high safety level in autonomous UAV operations it is crucial to have an accurate estimate of the vehicle's state at any given time.

Standard techniques deployed for pose estimation are Visual-Inertial Odometry (VIO) and Simultaneous Localization and Mapping (SLAM)~\cite{bloesch2015robust,mourikis2007multi,mur2017visual,engel2014lsd,klein2007parallel}. These methods fuse information from both visual and inertial sensors to localize the vehicle. 
Visual-only or inertial-only odometry estimators suffer from drifts and scale ambiguity. In visual odometry, loop closure can reduce the drift problem, but we still need to integrate external information to solve for the scale ambiguity. Therefore, by fusing inertial and visual data, not only do we resolve the scale ambiguity, but we also increase the accuracy of the odometry itself.

The pipeline for VIO and SLAM typically consists of camera calibration, followed by feature detection and tracking, outlier rejection, motion and scale estimation, optimization back-end, and local optimization (Bundle Adjustment). 
However, these techniques lack robustness when deployed in challenging conditions, such as low-texture or low-light environments, or in the presence of noises, blurs, camera occlusions, dynamic objects in the scene, and camera calibration errors ~\cite{han2010bio,bloesch2015robust,qin2018vins}.
Additionally, different scenarios may require different types of features for tracking and matching, as well as more adaptive algorithms. Nevertheless, VIO and SLAM methods only track pre-designed hand-engineered descriptors in all the contexts. 

The advent of deep learning technology has made neural networks more appealing for dealing with visual-inertial odometry problems~\cite{lecun2015deep}. 
By learning features from data rather than using hand-designed descriptors, deep neural networks can adapt to different contexts~\cite{clark2017vinet,kendall2015posenet}.


While neural networks can extract visual information about vehicle position and orientation, data gathered from inertial measurement units (IMUs)  provide complementary information about the pose. By combining neural networks that learn visual features with data gathered from inertial measurement units, we propose a multimodal fusion learning approach to increase odometer estimation accuracy

\subsection{Contributions}

We propose a new end-to-end approach for online pose estimation that leverages multimodal fusion learning. This consists of a convolutional neural network for image regression and two long short-term memories (LSTMs) of different sizes to account for both sequential and temporal relationships of the input data streams.
A small LSTM architecture integrates arrays of acceleration and angular velocity from the inertial measurements unit sensor. A bigger core LSTM processes visual and inertial feature representations along with the previous vehicle's pose and returns position and orientation estimates at any given time.
 
 We assess the performance of our model and compare it to a baseline algorithms for visual inertia odometry on the publicly available EuRoC MAV dataset. The results show that our method significantly outperforms the state-of-the-art for odometry estimation, improving the accuracy up to $25 \%$ over the baseline.
 
We then integrate our data-driven odometry module in a closed-loop flight control system, providing a new method for real-time autonomous navigation and landing. To this end, we generate a simulated \textit{Downtown} environment using Airsim, a flight simulator available as a plugin for Unreal Engine~\cite{airsim2017fsr}. We collect images and inertial measurements flying in the simulated environment and we train the model on the new synthetic dataset. The network outputs are now the input to the flight control system that generates velocity commands for the UAV system. We  show through real-time simulations that our closed-loop data-driven control system can successfully navigate and land the UAV on the designed target with less than $10$ cm of error.



 \begin{algorithm}[h!]

\SetAlgoLined
\KwIn{Input Image $I_t$,\\
\quad \quad \quad \quad IMU: $\omega_{t-T:t}, a_{t-T:t}$,\\
\quad \quad \quad \quad Previous Pose: $\xprev{x}$}
\KwOut{Estimated pose: $\xt{x} = \vector{x,y,q,q,q,q}$}
\While{Flying:}{
\For{$t=1,\ldots,T$}{
 Collect data (Images and IMU)\;
 \eIf{Camera Corrupted}{Use IMU only}{Use Image and IMU}
  1. $\Iv{y}$ = CNN($I_t$)\;
    2. $\Iimu{y}$ = LSTM($\omega_{t-T:t}, a_{t-T:t}$)\;
 3. $\zt{t}$ = concat($\Iv{y}, \Iimu{y}$)\;
 4. $\hatxt{x}$ = CoreLSTM($\zt{z}, \xprev{x}$)\
}}
\caption{Data-driven pose estimation}\label{Algorithm}
\end{algorithm}

\subsection{Previous Work}
Traditional localization approaches for autonomous UAV navigation rely on computer vision algorithms supplemented by sensors, including Global Positioning Systems (GPS) and Inertial Measurement Units (IMUs)~\cite{forster2014svo,carrillo2012combining,liu2017monocular,mourikis2007multi}.
Visual Servoing methods use the tracked features as inputs to a control law that directs the robot into a desired pose~\cite{fontanelli2009visual,hutchinson1996tutorial,han2010bio}.
However, environmental noise and the presence of moving objects can negatively affect the tracking process and reduce the estimation accuracy.
Additionally, VIO systems demand heavy computation due to image processing and sensor fusion. 
The advent of deep learning techniques has created new benchmarks in almost all areas of computer vision. In the last few years, CNN architectures for pose estimation have captured the interests of the robotics community ~\cite{wang2017deepvo,clark2017vinet,byravan2017se3,bateux2017visual}. PoseNet has been the first approach to use CNNs to address the metric localization problem~\cite{kendall2015posenet}. 
Additionally, other architectures have been deployed to estimate the incremental motion of the camera using only sequential camera images or a combination of visual and inertial data~\cite{wang2017deepvo,shakernia1999landing,herisse2011landing,eigen2014depth,clark2017vinet}.

\begin{figure*}[b!]
     \centering
      \framebox{\parbox{6in}{
     \includegraphics[width=\linewidth]{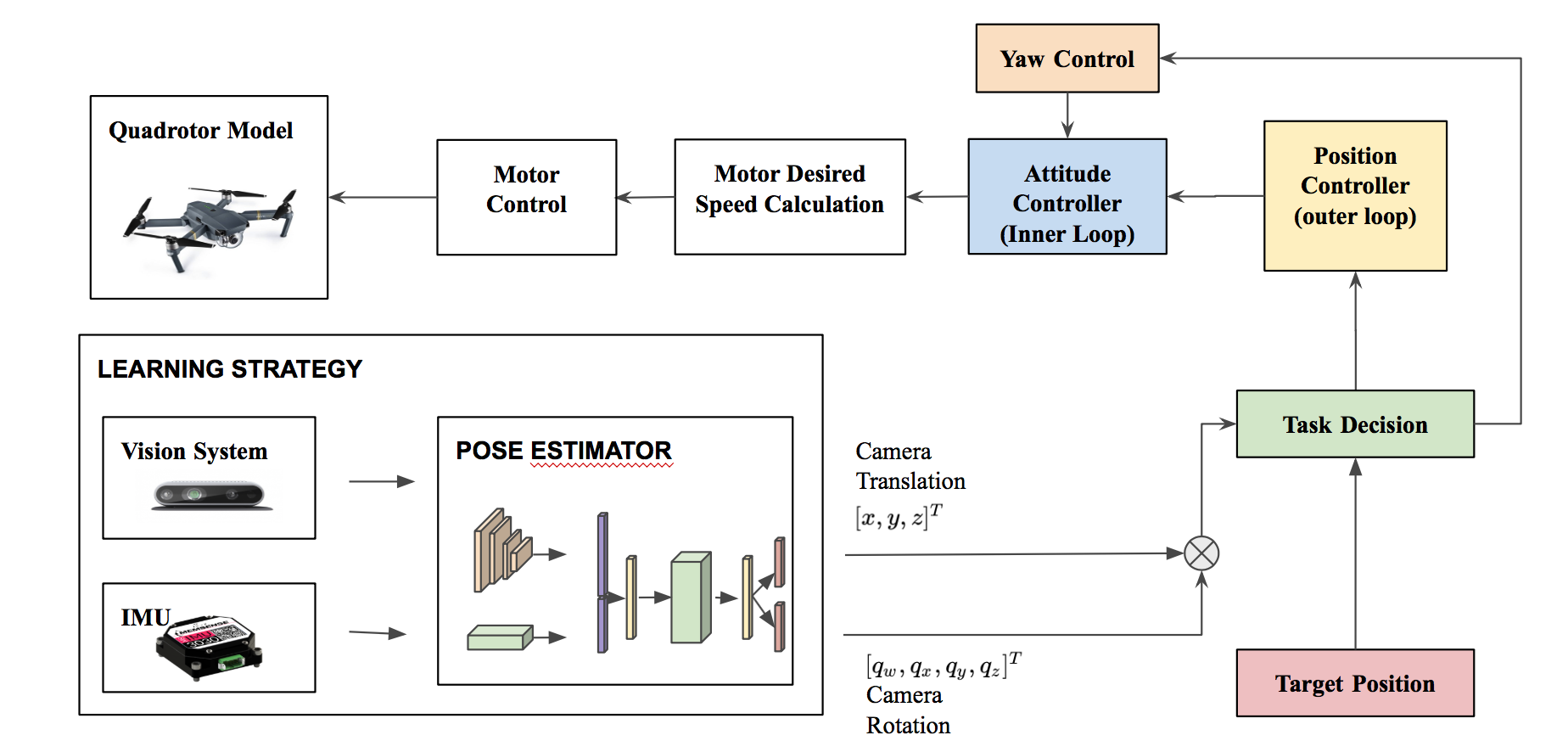}}}
      \caption{High-level illustration of the data-driven GNC model. The neural network makes a prediction about the robot position and orientation. The low level controller uses the predicted pose along with the desired one to generate velocity commands to drive the robot to its destination. }
      \label{control_scheme}
\end{figure*}

\section{PRELIMINARIES}\label{preliminaries}

\subsection{Localization Problem}
Given the actual state $ \xt{x} = \vector{x,y,q,q,q,q}\in \R^7$, we train the neural network to estimate the vehicle pose $\hatxt{x} = \hatvector{x,y,z,q,q,q,q} \in \R^7$ from a continuous stream of images and inertial measurements. The inputs for our model are observation tuples $\yt{y} =\{ \Iimu{y},\Iv{y} \} $ of RGB images ($\Iv{y}$) and IMU data ($\Iimu{y}$), where $\Iimu{y} = \imuvector{tau,a,a,a,\omega, \omega, \omega} \in \R^{N \times 7}$, $\tau_t$ is the timestamp of the inertial measurement, $\yt{s}$ is the linear acceleration, $\yt{\omega}$ is the angular velocity, and $N$ is the number of inertial observation between two consecutive camera frames $t$ and $t+1$.
The online localization task aims to estimate the pose of the vehicle $\xt{x}$ at any given time given the current observations $\yt{y}$ and previous pose state $\xprev{x}$. In the learning framework, we aim to model the mapping $f$ between raw data and the current pose as follows: $\xt{x} = f(\xprev{x}, \yt{y})$, $f:\R^6, \R^{p \times q} \rightarrow \R^{7}$, where $p,q$ are the image dimensions.

\subsection{Control Problem}
We make the assumption that the target position $\xdespos{x}=\desvec{x,y,z}$ is know.
After the controller receives the reference position of the target $\xdespos{x}$, the desired velocities $\xdesvel{x} =\desvec{u,v,w}$ are computed based on the rate of change of position set points, i.e., $\xdesvel{x} = \dotxdespos{x}$.
We use the velocity reference $\xdesvel{x}$ along with the position reference $\xdespos{x}$ to compute the final throttle and attitude angle commands $\ucom{u} = \ucomvec{F,\psi,\theta,\phi}$ that are then fed back into the low-level controller. 
Given the target pose coordinate, we simulate a control law for $\ucom{u} $ depending only on $\xdespos{x}$ and $\dotxdespos{x}$ such that $\xt{x} \rightarrow \xdespos{x}$  and $ \vomega{v,\omega} \rightarrow 0$.
Fig.~\ref{control_scheme} illustrates the control system architecture.


\section{Architecture}\label{Architecture_sec}

Fig.~\ref{architecture} depicts the architecture of our model.
The inputs to the network are synchronized data from visual and inertial sensors.
We estimate the UAV's absolute pose by minimizing the geometry consistency loss function described in \ref{sec:loss}.

We train the network in an end-to-end fashion to regress the vehicle pose from sequences of images and windows of inertial measurements collected between two consecutive image frames. 

\paragraph{Image feature extractor}
To encode image features, we use ResNet18, pre-trained on the ImageNet dataset, truncated before the last average pooling layer. Each of the convolutions is followed by batch normalization and the Rectified Linear Unit (ReLU).
We replace the average pooling with global average pooling and subsequently add two inner-product layers. The output is a visual feature vector representation $z_{V}$.

\paragraph{Inertial feature extraction}
IMU measurements are generally available at a rate of an order of magnitude higher (e.g., $~100-200 Hz$) than visual data (e.g., $~10-20 Hz$).
A Long Short-Term Memory (LSTM) processes batches of IMU data (acceleration and angular velocity) between two consecutive image frames and outputs an inertial feature vector $\Iimu{z}$.
LSTM exploits the temporal dependencies of the input data by maintaining hidden states throughout the window.

\paragraph{Intermediate fully-connected layer}
The inertial feature vector $\Iimu{z}$ is concatenated with the visual feature representation $\Iv{z}$ into a single feature $z_t$ representing the motion dynamics of the robot: $\xt{z} = \mathtt{concat}(\Iv{z}, \Iv{I})$.
This vector is then carried over to the core LSTM for sequential modeling.

\paragraph{Core LSTM}
The core LSTM takes as input the motion feature $z_t$ along with its previous hidden states $\hprev{h}$ and models the dynamics and the connections between sequences of features, where  $\zt{h}= \mathit{f}(\zt{z},\hprev{h})$.  The use of the LSTM module allows for the rapid deployment of visual-inertial pose tracking.
These models can maintain the memory of the hidden states over time and have feedback loops among them. In this way, they enable their hidden state to be related to the previous one, allowing them to learn the connection between the last input and pose state in the sequence.
Finally, the output of the LSTM is carried into a fully-connected layer, which serves as an odometry estimation. The first inner-product layer is of dimension $1024$, and the following two are of dimensions $3$ and $4$ for regressing the translation $x$ and rotation $q$ as quaternions. Overall, the fully connected layer maps the features vector representation $\zt{z}$ into a pose vector as follows:  $\xt{x} = LSTM(\zt{z}, \hprev{h})$.

\subsection{Learning and Inference}\label{sec:loss}

To regress the pose of the vehicle, we compute the Euclidean loss between the estimated pose and the ground truth. We adopt Adam optimizer to minimize this loss function, starting with an initial rate of  $10^{-4}$~\cite{kingma2014adam}.

 \begin{figure}[h!]
      \centering
      \framebox{\parbox{3in}{
     \includegraphics[width=\linewidth]{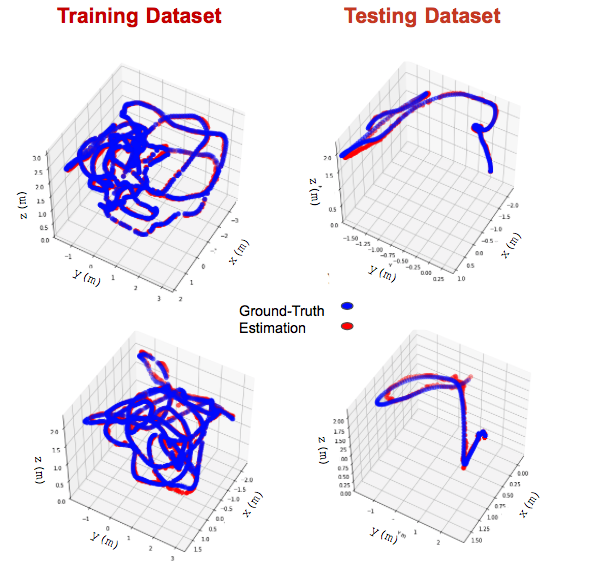}
}}
      \caption{Sampled 3D trajectories of results on the training and testing sequences }
      \label{traj}
  \end{figure}

\paragraph{Loss Function}
We predict the position and orientation of the robot following the work of Kendall et al., with the following modification~\cite{kendall2017geometric}. In our loss function, we introduce an additional constraint that penalizes both the $L_1$ and $L_2$ Euclidean norm. Let $\xt{x}= \xyzvec{x,y,z} \in \R^3, \qb{q} = \qvec{q,q,q,q} \in \R^4$ be the ground-truth translation and rotation vector, respectively, and $\hat{\xt{x}}, \hat{\qb{q}}$ their estimates. 

Our loss function is as follows: \begin{align}
    \mathcal{L}_{\beta}(I)=\mathcal{L}_{x}(I)+\beta \mathcal{L}_{q}(I)
\end{align}
 where $$\mathcal{L}_{x}(I)=\|\hat{\xt{x}}-\xt{x}\|_{L_2} + \gamma \|\hat{\xt{x}}-\xt{x}\|_{L_1} $$  and
 $$\mathcal{L}_{q}(I)=\left\|\hat{\qb{q}}-\frac{\qb{q}}{\|\qb{q}\|}\right\|_{L_2} + \gamma \left\|\hat{\qb{q}}-\frac{\qb{q}}{\|\qb{q}\|}\right\|_{L_1}$$ represents the translation and the rotation loss. 
$\beta$ is a scale factor that balances the weights of position and orientation, which are expressed in different units, and $\gamma$ is a coefficient introduced to balance the two Euclidean norms. 
However, $\beta$ requires significant tuning to get consistent results, as shown in ~\cite{kendall2017geometric}. To avoid this issue, we replace $\beta$ by introducing learnable parameters.
The final loss function is as follows: \begin{align}
    \mathcal{L}_{\sigma}(I)=\mathcal{L}_{x}(I) \exp \left(-\hat{s}_{x}\right)+\hat{s}_{x}+\mathcal{L}_{q}(I) \exp \left(-\hat{s}_{q}\right)+\hat{s}_{q}
\end{align}
where $\hat{s} :=\log \hat{\sigma}^{2}$ is the learnable variable and each variable acts as a weight for the respective component in the loss function.

\begin{figure}[h!]
      \centering
     \includegraphics[width=\linewidth]{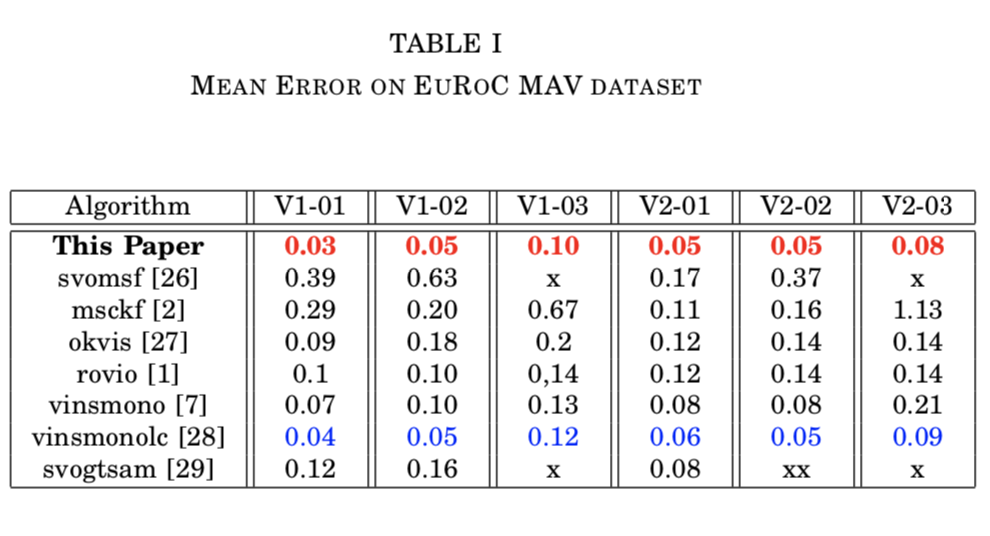}
     \label{fig:table}
  \end{figure} 

\section{BENCHMARKING}\label{Experiments}

All the experiments are carried out on an Intel Xeon CPU E5-2650 @ 2 GHz processor 192 GB RAM, and NVIDIA TITAN RTX GPU.
We first evaluate the performance of the estimator by comparing our results to the state-of-the-art for VIO on the EuRoC MAV dataset provided by \cite{delmerico2018benchmark}. This dataset consists of eleven visual-inertial challenge sequences recorded onboard a micro-aerial-vehicle (MAV) flying in an indoor environment. It provides stereo monochrome images at $20$ Hz, temporally synchronized IMU data at $200$ Hz, and ground-truth positioning measurements from the  Vicon motion capture system.

For quantitative pose evaluation, we compute the average root mean square (RMS) translation and rotation error and compare our results to traditional methods reported in \cite{delmerico2018benchmark,kaiser2016simultaneous,mourikis2007multi,leutenegger2013keyframe,bloesch2015robust, qin2018vins,lynen2013robust,forster2016manifold} .
The translation errors are computed across the entire length of the trajectory and ground truth. The RMSE for regression represents the sample standard deviation of the differences between predicted values and real values 
\begin{align}
R M S E=\sqrt{\frac{\sum_{i=1}^{I}\left(\hat{y}_{i}-y_{i}\right)^{2}}{n}}
\end{align}{}
where $y$ is the real and $\hat{y}$ is the predicted one.

Table I shows a comparative analysis of average translation RMSE obtained with our method and other traditional odometry estimators on the EuRoC MAV dataset.
This dataset contains sequences of videos with different characteristics, some of which are more challenging than others.
In all the experiments, our method outperforms the baseline.

 \begin{figure}[h!]
\centering
 \framebox{\parbox{3.3in}{
     \includegraphics[width=\linewidth]{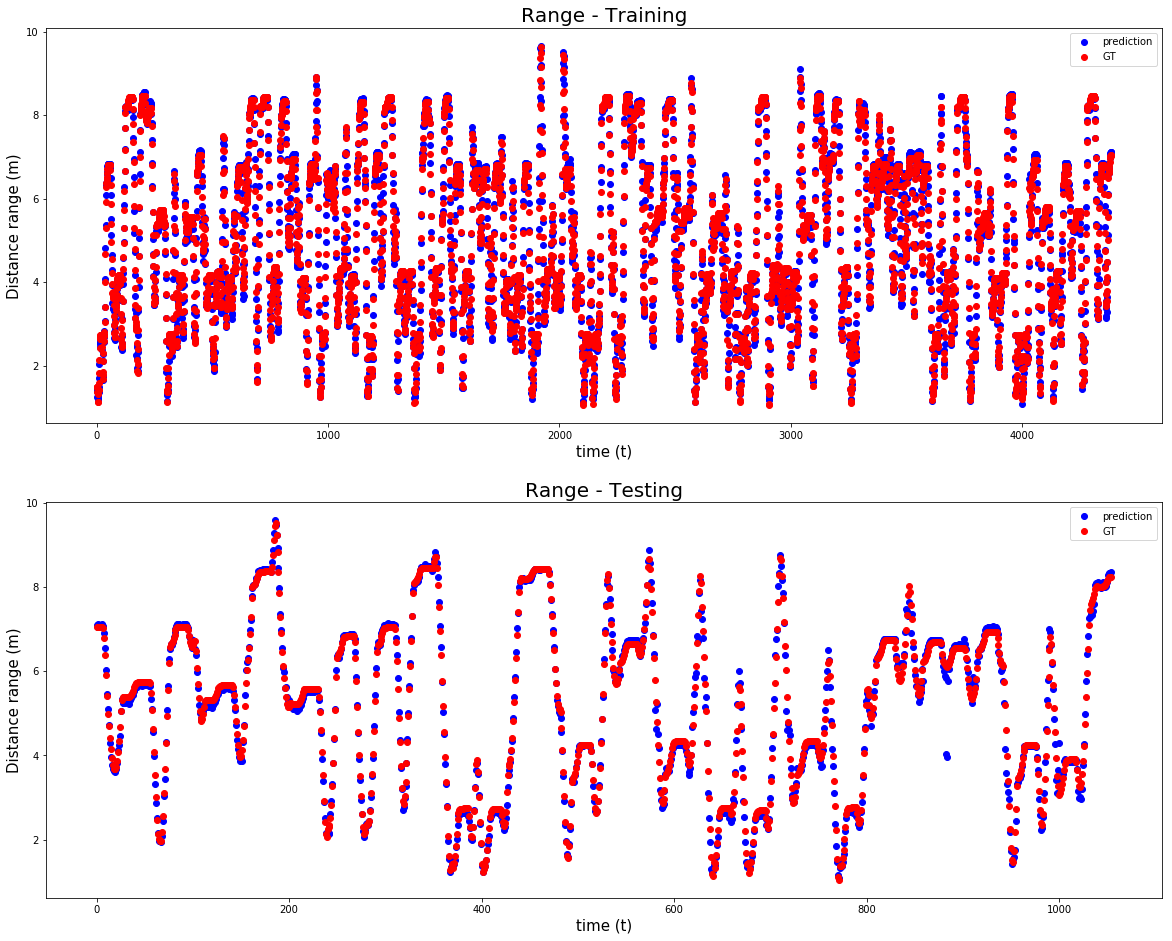}
}}
      \caption{Range measurements for training and testing dataset}
      \label{traje}
\end{figure}

\section{INTEGRATION WITH THE FLIGHT CONTROL SYSTEM}

\subsection{The Dataset}
\begin{figure}[h!]
      \centering
      \framebox{\parbox{3.3in}{
            \centering
     \includegraphics[width=0.46\linewidth]{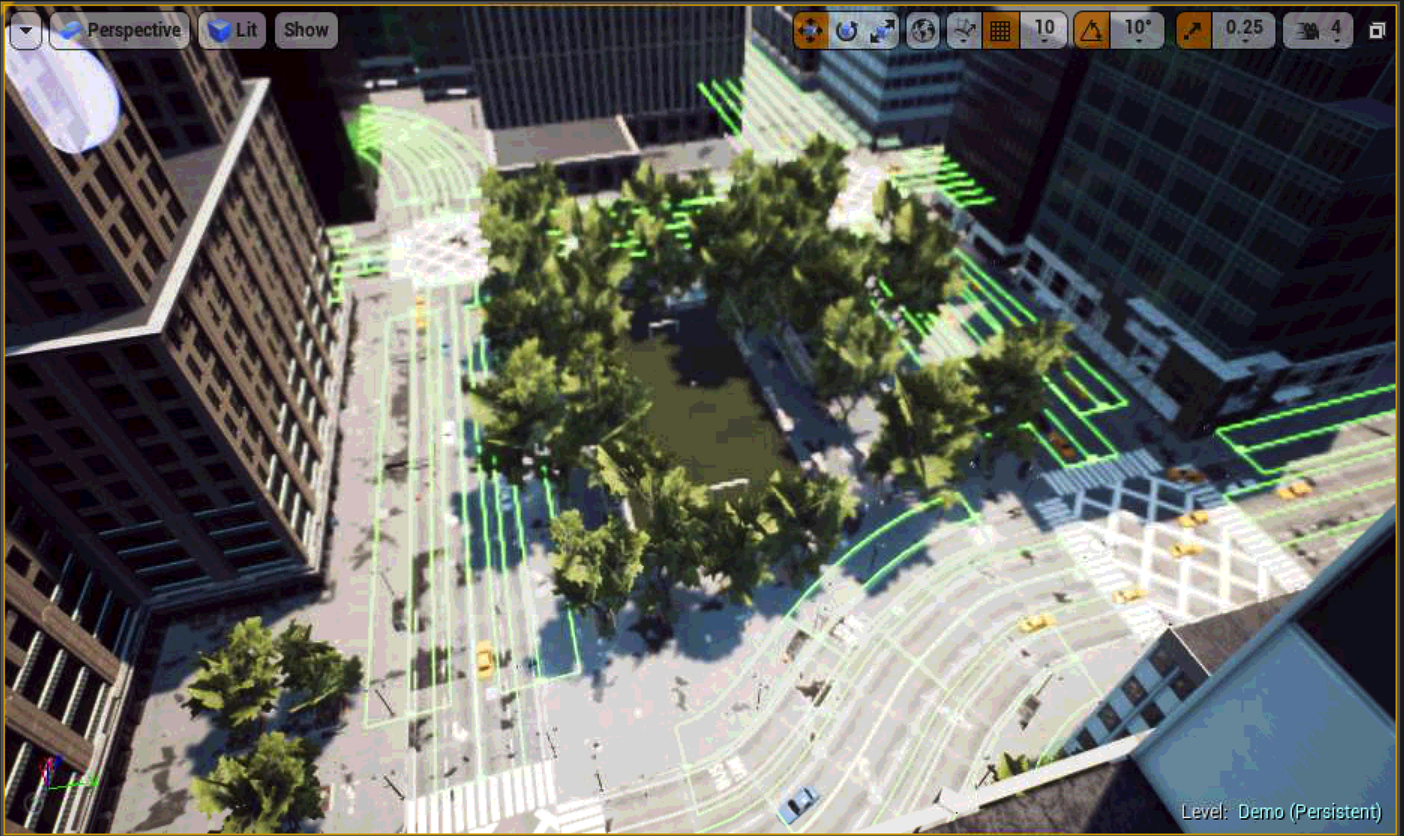}
        \includegraphics[width=0.51\linewidth]{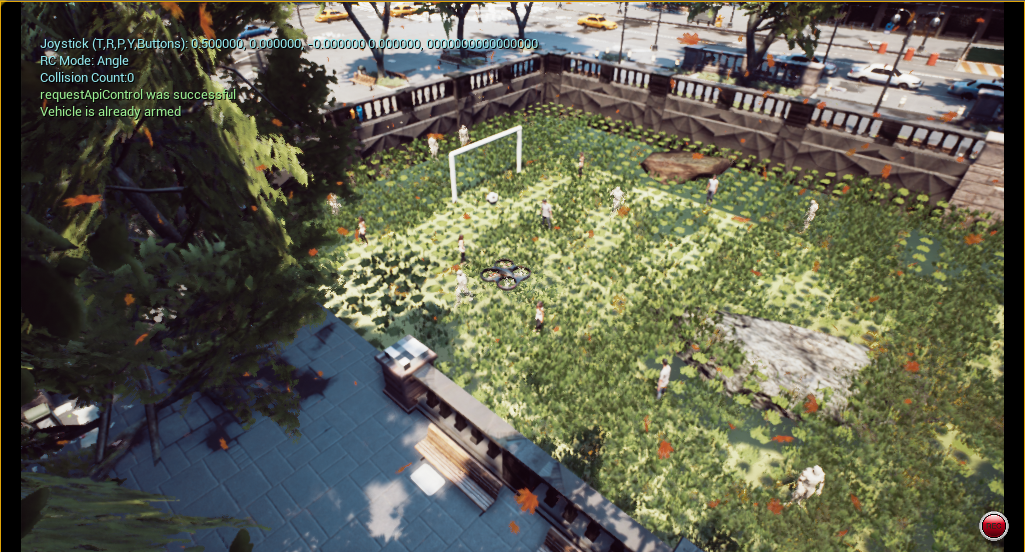}
}}
      \caption{Images sampled from the \textit{Downtown} simulated environments used in the paper.}
      \label{downtown1}
   \end{figure}

To generate the testing scenario, we use Airsim~\cite{airsim2017fsr}, an open-source simulator that aims to close the gap between simulation and reality.
As a plugin, we can use Airsim in any environment developed for Unreal Engine (UE).
We collect training data in the virtual \textit{Downtown} environment retrieved from the Unreal Engine marketplace.
Fig.~\ref{downtown1} shows an example scenario.
We split each dataset into two sub-datasets, one for training and the other for testing.
Each flight within the dataset contains timestamped values for the ground-truth $6$ DOF pose of the UAV at $100 Hz$, IMU measurements at $100 Hz$, and camera streams (downward-facing) at $10 Hz$.  Ground-truth and sensor data are then pre-processed and synchronized.

\subsection{Experiment Implementation}

In the simulation, the UAV collects training data flying at a constant velocity over a grass field, landing on top of randomly chosen pillars. The UAV is equipped with a proportional-integral-derivative (PID) flight controller (FC) that maintains fixed altitude and takes as input the current pose of the vehicle.
We label the recorded frames of the simulator with the corresponding pose measurements from the dataset. 

The total simulation time is $5 h$, and we collect images with a step-size of $0.1$ seconds. We then split the dataset into two sub-datasets with a ratio of $0.8$ and $0.2$ as a training and testing set, respectively. We finally downsample the resolution of images to $512 \times 288 \times 3 $ (RGB) to reduce the computational cost. Each image is normalized to the range $[0,1]$. Both datasets present corrupted data images. To deal with this issue, we trained the network to use only IMU data as a new corrupted imaged shows up, as shown in Algorithm~\ref{Algorithm}.

\begin{figure}[h!]
\centering
 \framebox{\parbox{3.3in}{
 \centering
                  \includegraphics[width=\linewidth]{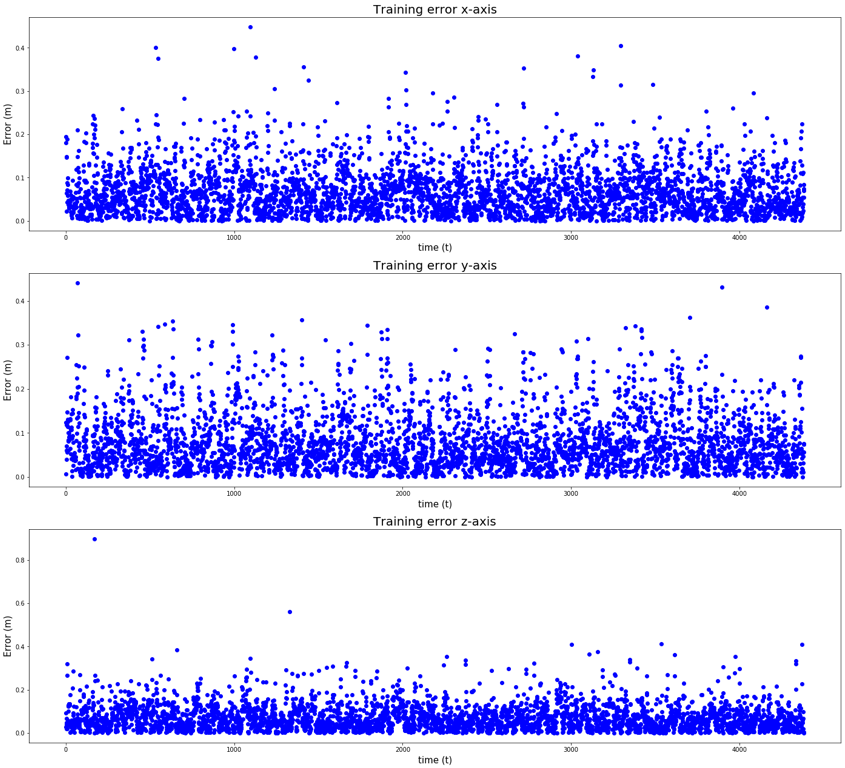}\\
                    \includegraphics[width=\linewidth]{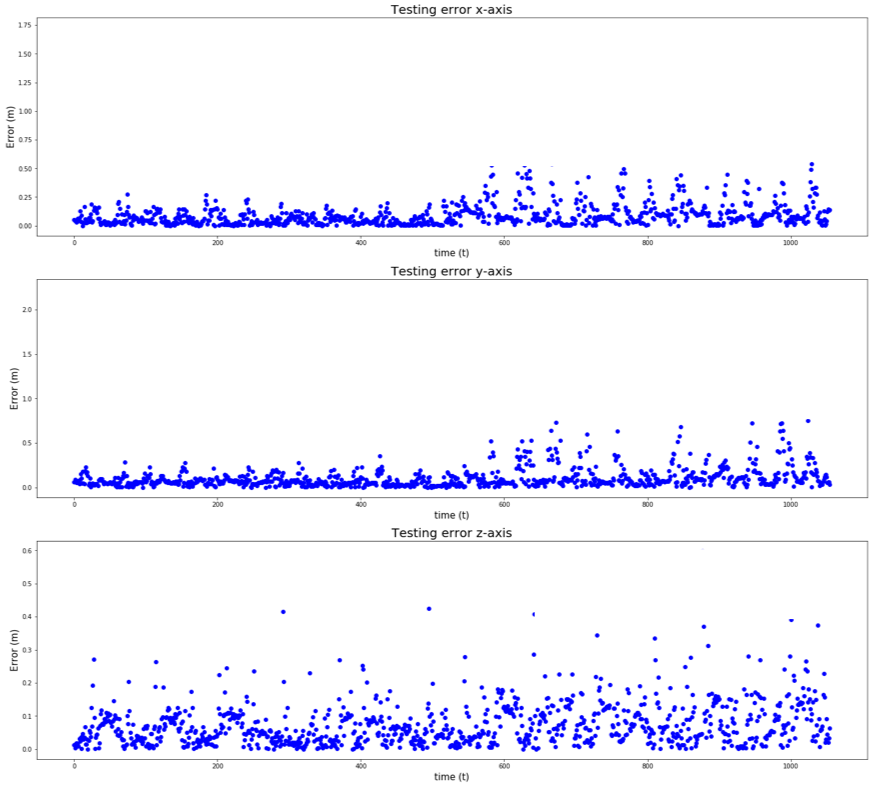}
}}
      \caption{Localization RMSE for training and test dataset}
      \label{trajerr}
\end{figure}

\paragraph{Lower bounds from steady-state Kalman filter}
 
Kalman filters (KF) are the optimal state estimator for systems with a linear process and measurement models.
To quantify the accuracy of our estimation, we use a worst-case estimation error derived from the steady-state covariance of the KF. This error is used to define a lower bound on the neural network performance.  

We consider the following discrete-time linear Gaussian state-space model  consider the following discrete-time linear Gaussian state-space model:
\begin{align}
x_{t+1} &=A \xt{x}+w_t,\quad  t \in \N\\
y_t &=H \xt{x} + \nu_t,
\end{align}{}
 where $\boldsymbol{x_t}$ is the state of the system and $\boldsymbol{y}_t$ is the measurements vector. $A$ and $H$ are matrices of appropriate dimensions. The process noise and the measurement noise are distributed according to $w_t \approx \mathcal{N}(0,Q)$ and $\nu_t \approx \mathcal{N}(0,R) $ respectively, with $Q,R$ covariance matrices.
 This model can be used in a target-tracking context to describe a linear target motion and measurement model. 
We assume that the sensor can measure only the position of the target:
\begin{align}\label{eq:distance}
    d_{t}=\sqrt{\left(x_{t}-x_{t}\right)^{2}+\left(y_{t}-y_{t}\right)^{2}+\left(z_{t}-z_{t}\right)^{2}}+\tilde{d}_{t}
\end{align}
where $\tilde{d}_{t}$ is the error in the measurement (+/- 1 pixel).

    \begin{figure}[h!]
      \centering
      \framebox{\parbox{3.3in}{
     \includegraphics[width=\linewidth]{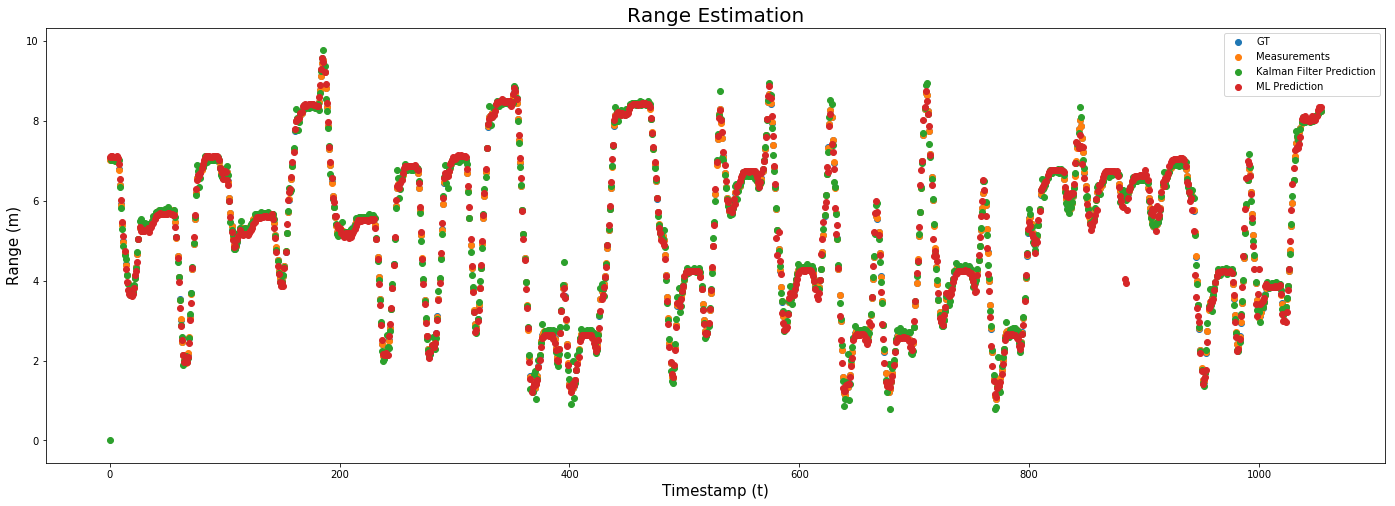}\\
          \includegraphics[width=\linewidth]{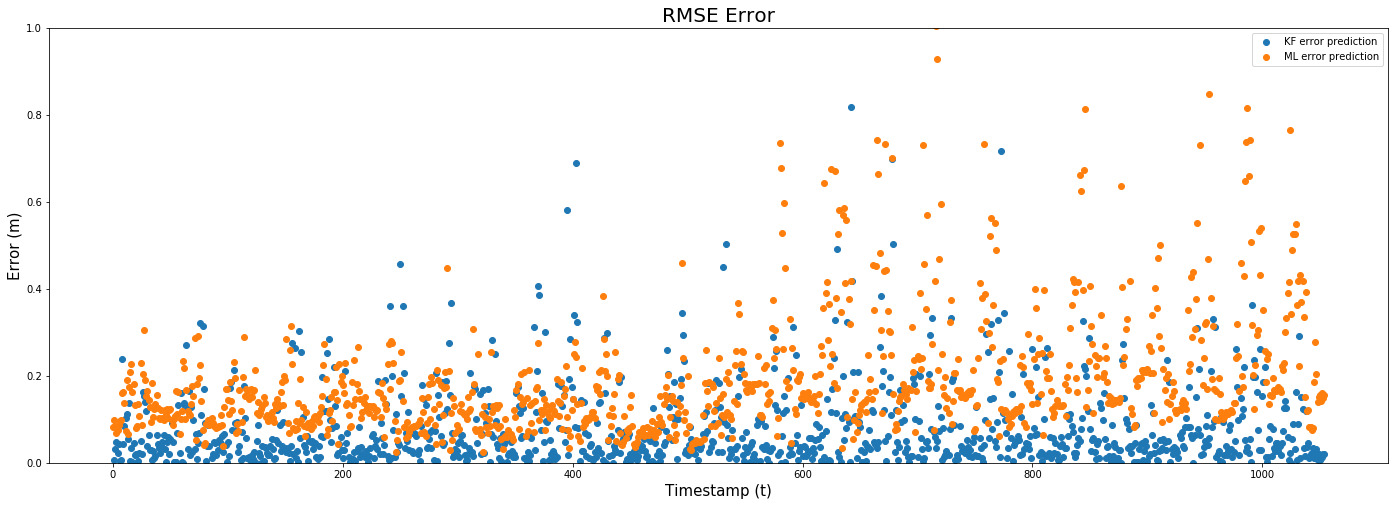}
}}
      \caption{ Kalman Filter (blue) and ML (orange) tracking prediction and RMSE.}
      \label{fig:kferror}
   \end{figure}
  \FloatBarrier

The one step ahead prediction for the performance of a Kalman
Filter is given by $P_{t | t}=\left(I- K_{t} H\right) P_{t | t-1}$.
We can now formulate Kalman-like recursions for an overall system as follows:\begin{align}
    \begin{aligned}
    P_{t+1 | t} &=A P_{t | t} A^{T}+Q \\ 
    K_{t} &=P_{t | t-1} H^{T}\left(H P_{t | t-1} H^{T}+R\right)^{-1} \\
    P_{t | t} &=\left(I- K_{t} H\right) P_{t | t-1} \end{aligned}
\end{align}

By performing the proper substitutions, we obtain \begin{align}\label{riccatiew}
P_{t+1 | t} &= A P_{t | t-1} A^{T}- A P_{t | t-1} H^{T}\left(H P_{t | t-1} H^{T}+R\right)^{-1}\\\nonumber
 &\times  H P_{t | t-1} A^{T}+Q
\end{align}{}
which represents the standard Riccati difference equation associated with the Kalman filter.
For $k \rightarrow \infty$, it has a steady-state algebraic equivalence given by \begin{align}
P &=A P A^{T}- A P H^{T}\left(H P H^{T}+R\right)^{-1} H P A^{T}+Q
\end{align}{}
In general, it is difficult to compare Kalman filters with neural networks. The Kalman filter utilizes a linear system for the localization estimates, whereas neural networks localize the vehicle by learning the mapping between sensor data and $6$-DOF poses. However, it is possible to use the steady-state covariance from a Kalman filter as a quality measure for the learning-based estimation.
Based on the experimental results, we found an empirical bound for the ML estimation equal to $10 cm$.
Fig.~\ref{fig:kferror} shows the RMSE error comparison between the KF, our estimate, and the ground-truth.

\begin{figure}[h!]
\centering
      \framebox{\parbox{3in}{
      \centering
       \centering
       \includegraphics[width=0.8\linewidth]{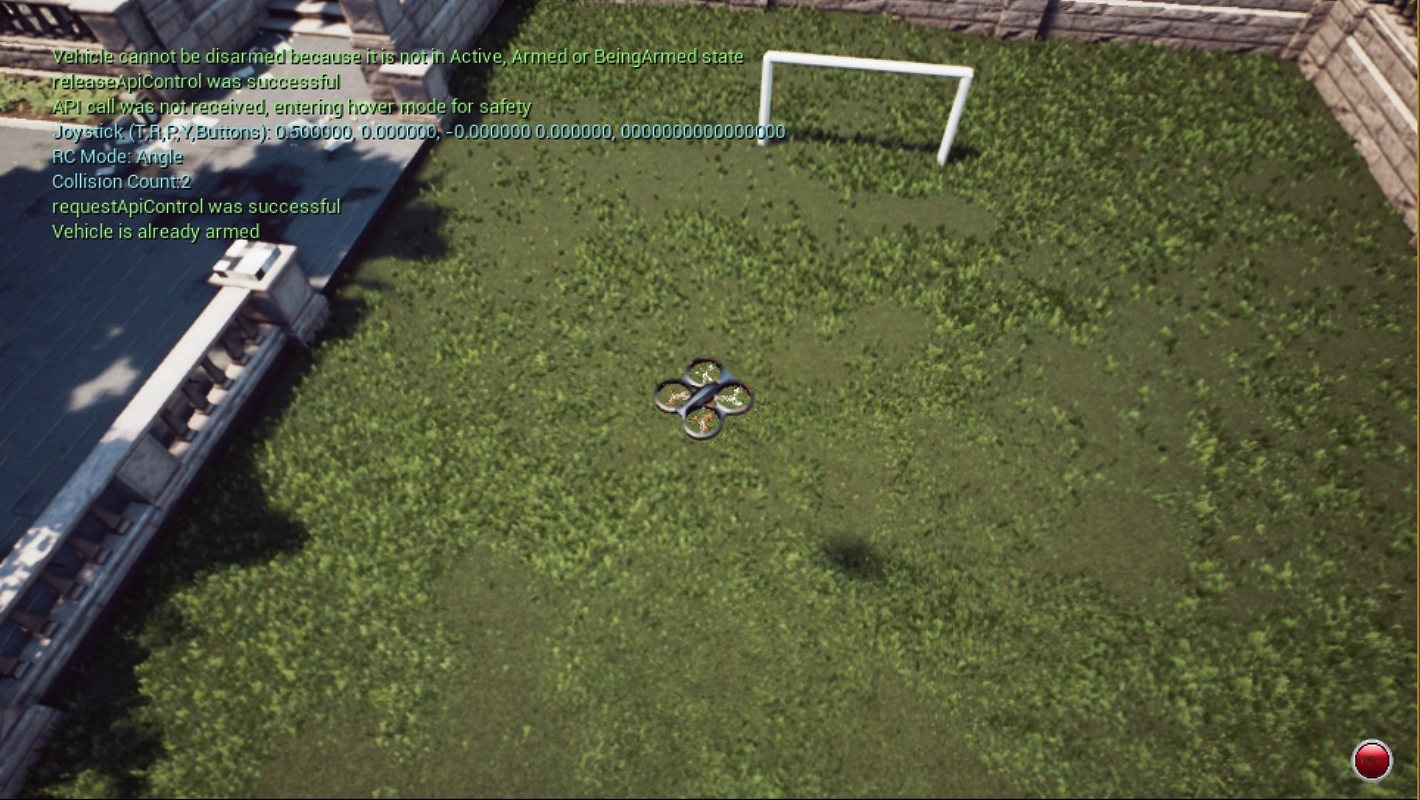}\\
        \vspace{0.2em}
 \vspace{0.04em}
       \centering
       \includegraphics[width=0.8\linewidth]{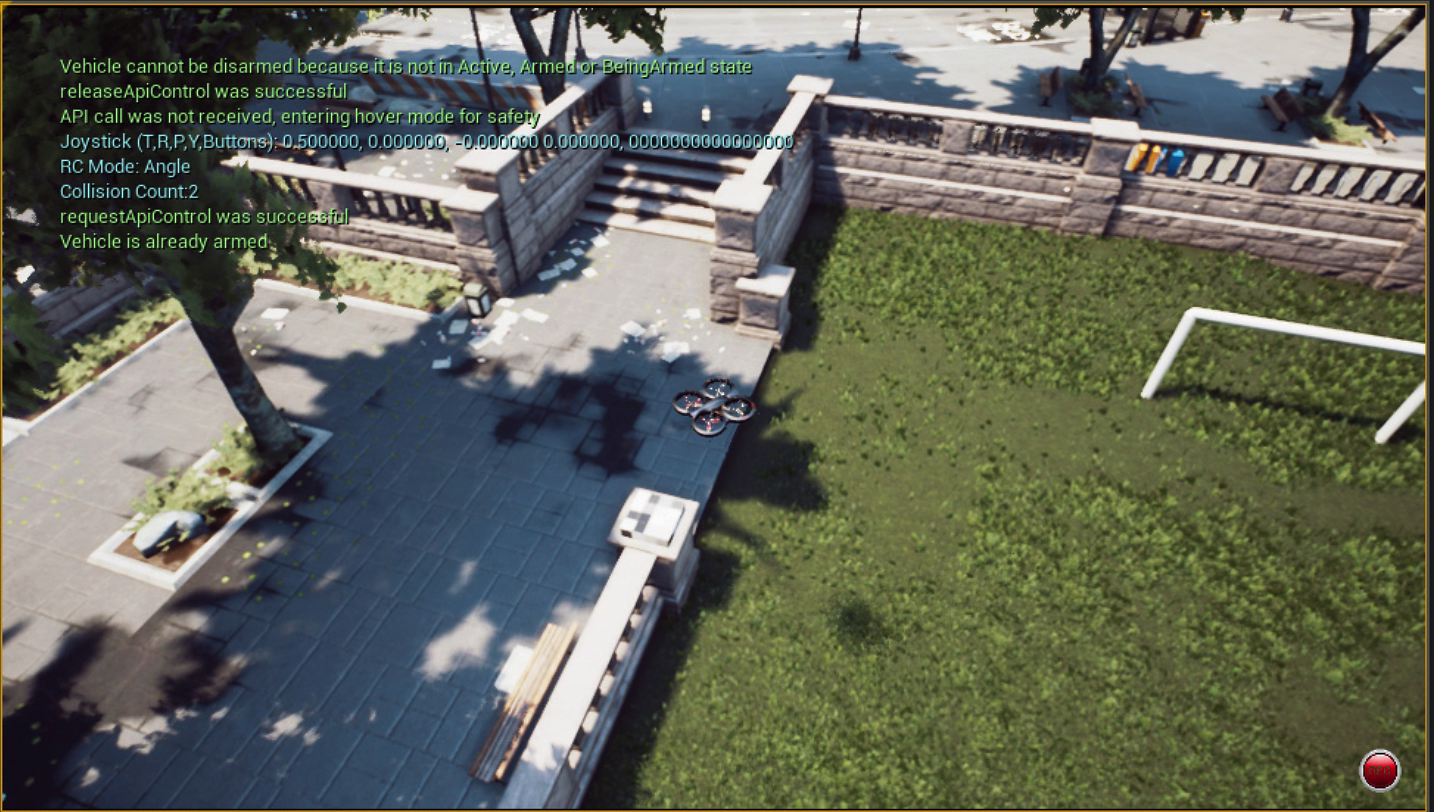}\\
               \vspace{0.2em}
     \vspace{0.04em}
       \centering
      \includegraphics[width=0.8\linewidth]{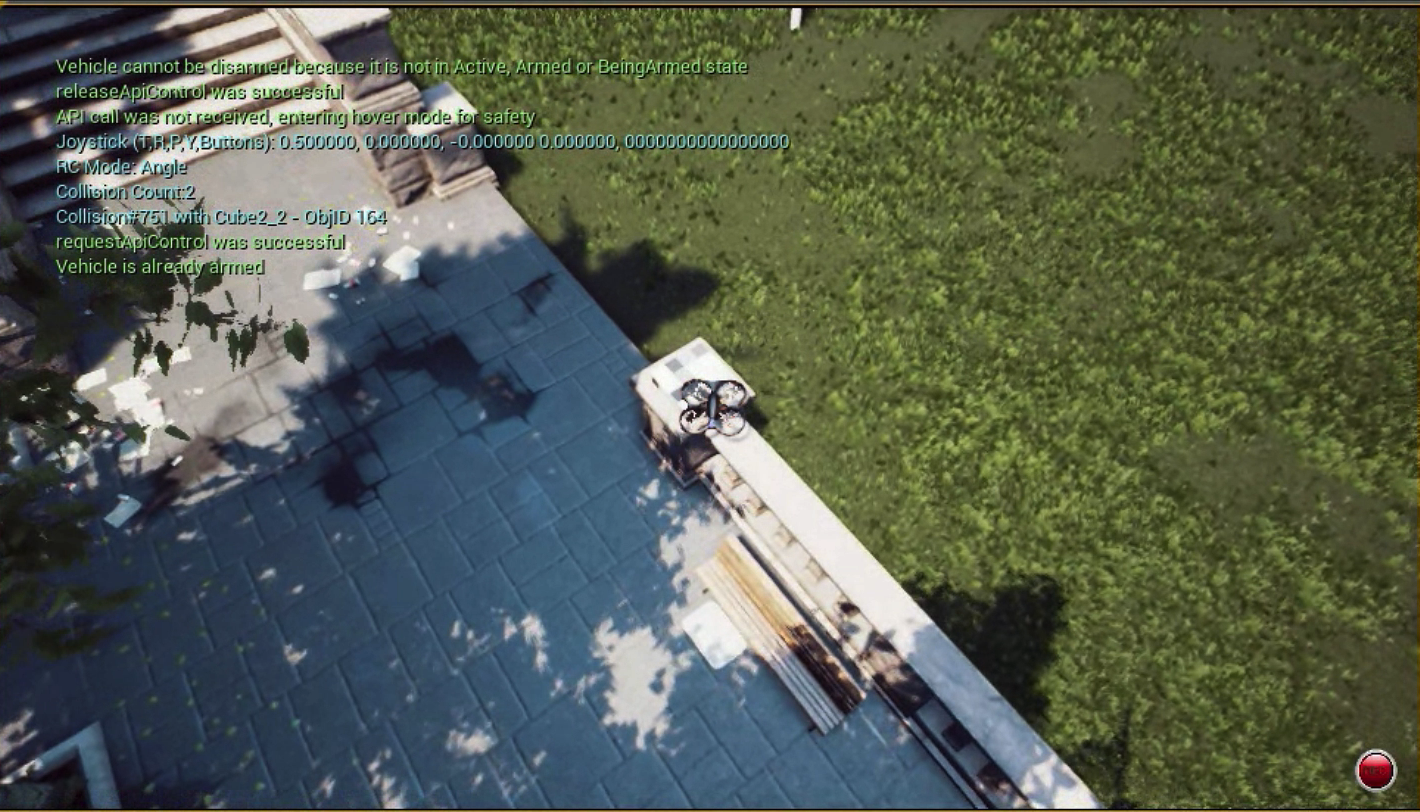}\\
 \includegraphics[width=0.49\linewidth]{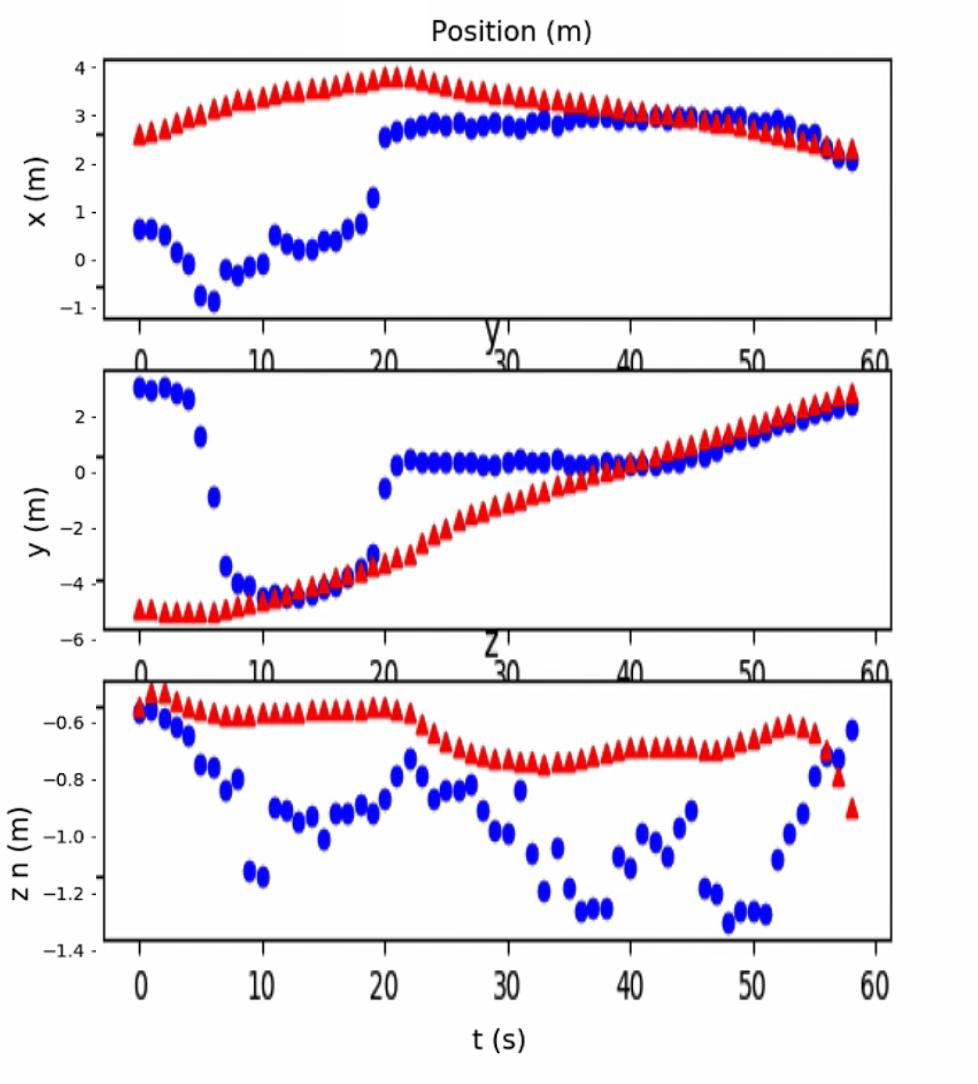}
         \includegraphics[width=0.48\linewidth]{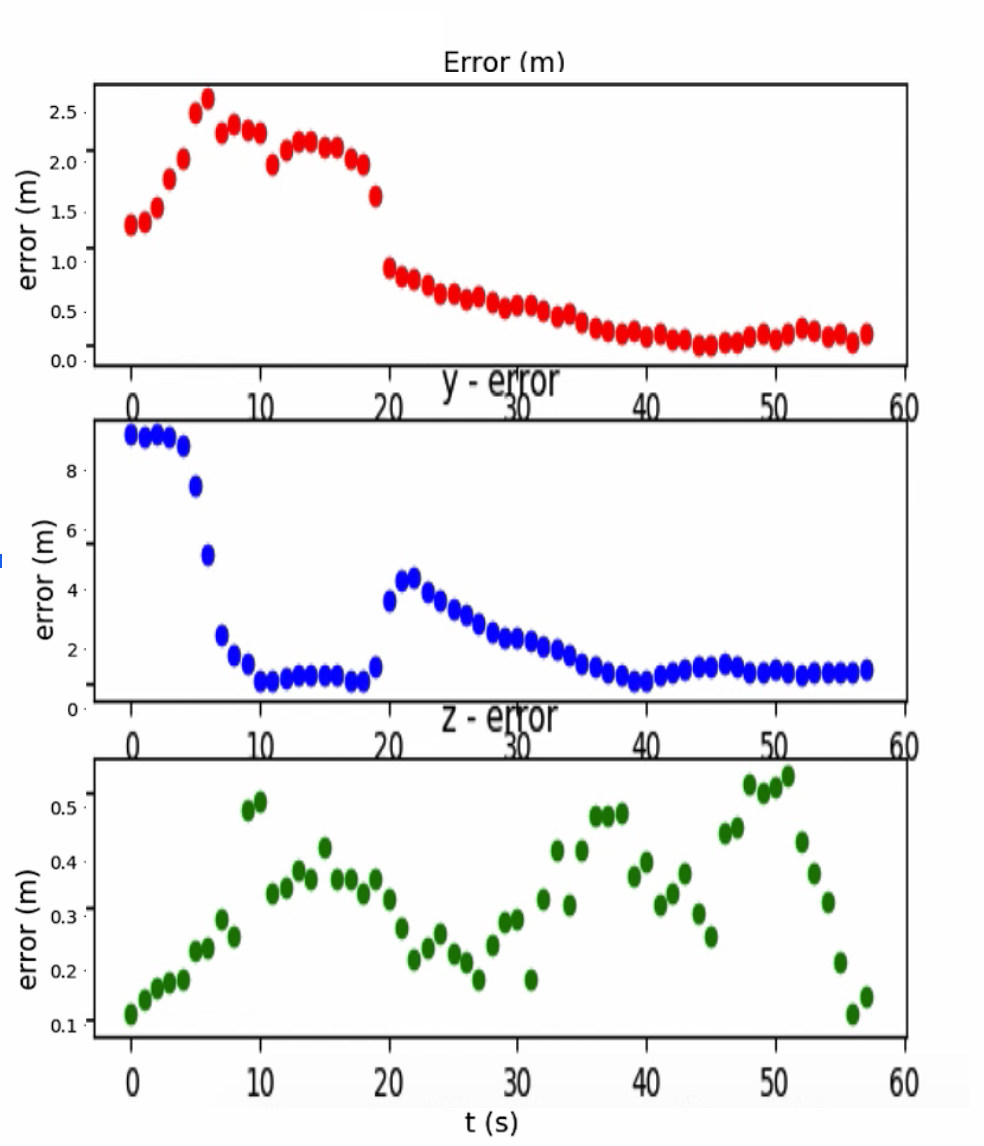}\\
    }}
\caption{UAV autonomous navigation and landing in AirSim.}
\label{landing}
\end{figure}

\subsection{UAV Landing}

\begin{figure}[h!]
     \centering
      \framebox{\parbox{3.3in}{
     \includegraphics[width=\linewidth]{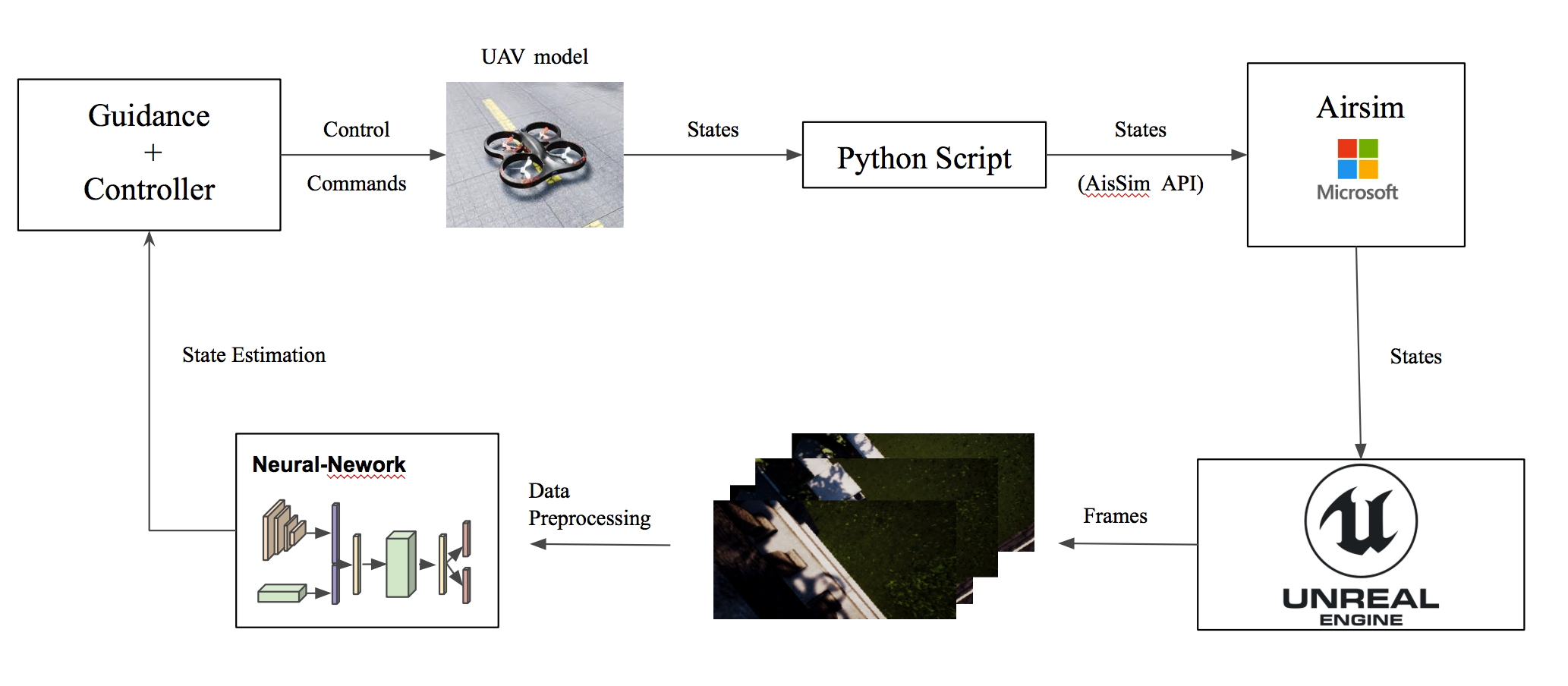}}}
      \caption{Neural Feedback Control Framework}
      \label{neural_control_scheme}
\end{figure}

We finally integrate the neural network for pose estimation in a closed-loop flight control system, and we use Airsim as a default simulator. Fig.~\ref{neural_control_scheme} exploits the general framework of the integration.
Airsim has a built-in Python API.  A non-custom Python script receives state estimated by our model and sends these inputs to the Unreal environment through the Airsim API. 
We capture asynchronous images and inertial measurements with associated labels, i.e., real state information. 
The guidance logic takes as input the current estimate along with the desired target pose and outputs reference velocities to the low-level controller. This input command is then used to control the motors of the vehicle.
Fig.~\ref{landing} illustrates an example of the simulated UAV landing.
During the take-off phase, the model is not able to localize the robot accurately because of a lack of training data in that specific area. However, by using information from the inertial measurements, we are finally able to converge to accurate positioning of the vehicle and to perform the landing on the desired target.
Fig.~\ref{neural_control_scheme} illustrates the system architecture for autonomous landing in Airsim and Unreal Engine.

\FloatBarrier

\subsection{Comparison to feature-based methods}

 Traditional approaches for VIO use feature descriptors such as SIFT, SURF, and ORB~\cite{karami2017image} to detect distinguishable points in the image, such as corners and edges.  These features, however, are chosen according to an engineer's judgment and a long trial and error process. 
To localize the vehicle, feature-based VIO systems are required to track a significant number of features through consecutive images, failing when this requirement is not met.
However, given the low-frequency operation rate of the camera, we might not be able to receive temporally-close sequences of images, leading to an unsuccessful initialization or loose of the track.
\begin{figure}[h!]
     \centering
      \framebox{\parbox{3.3in}{
      \centering
     \includegraphics[width=\linewidth]{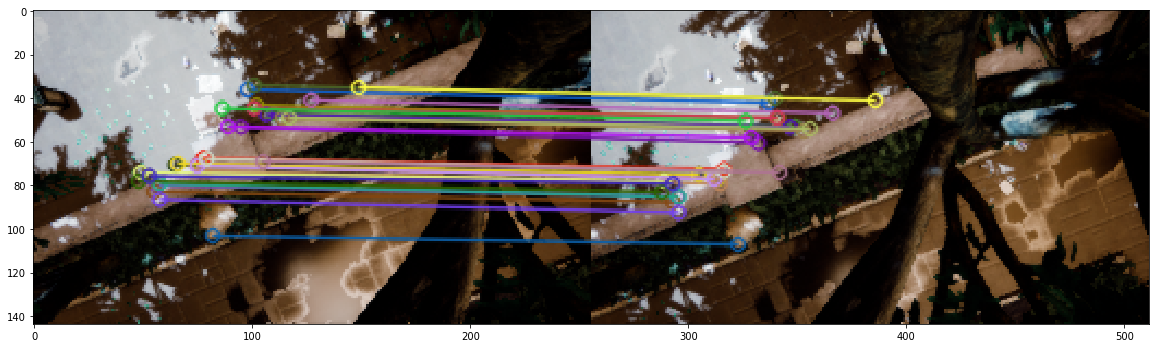}\\
          \includegraphics[width=\linewidth]{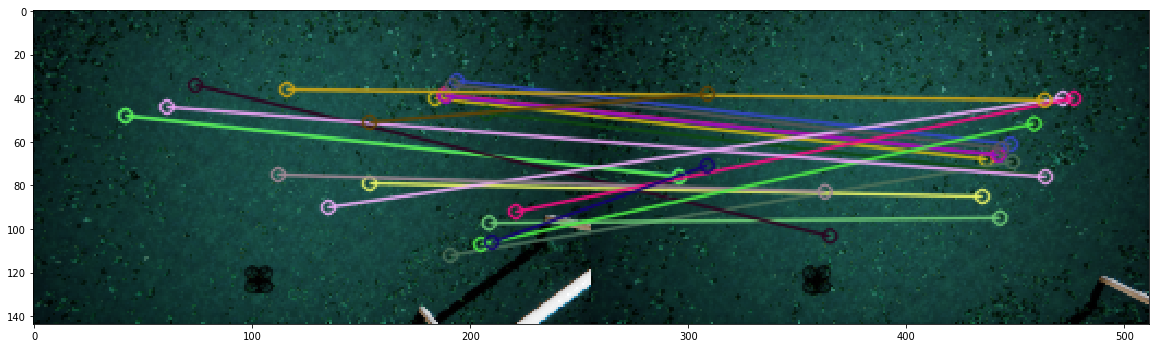}}}
     \caption{a) Good vs bad matches. In the first case, the features are easily recognizable, and the algorithm performs a good match between two consecutive frames. In the second case, the wind and the light changes make the tracked points not recognizable in the following frame yielding to a features mismatch.} 
      \label{matches1}
\end{figure}
In contrast, learned features are more effective than hand-engineered ones. CNNs are trained to learn features  rather than programmed. Hence, CNNs can learn more descriptive and salient features compared to traditional methods, developing better representations for the image data.
For this reason, deep learning methods outperform traditional feature-based algorithms.

\section{DISCUSSION AND FUTURE WORK}\label{Conclusion}

The advantage of using deep neural networks over traditional methods relies on its capability to produce a state estimate in just a single step.
However, one drawback of supervised deep learning networks is their requirement of abundant training data that must be sampled across all expected operating conditions. If there is not enough training data to present all the expected UAV operating conditions, then the network may not be able to perform accurate localization, as shown in Fig.~\ref{landing}.  

\paragraph{Learning to Generalize to Dynamic Environments with Meta-Learning}

\begin{figure}[h!]
      \centering
      \framebox{\parbox{3in}{

               \includegraphics[width=0.51\linewidth]{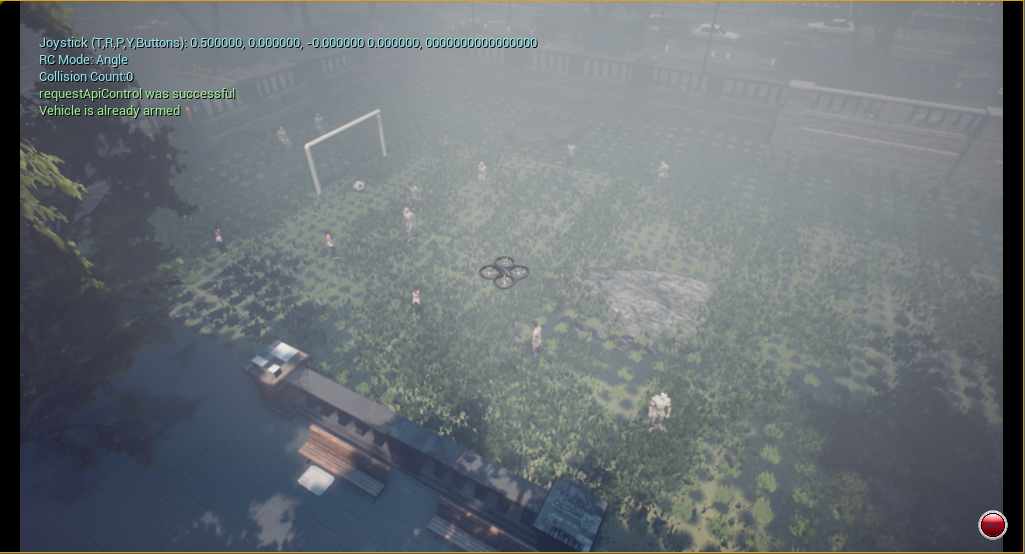}
          \includegraphics[width=0.47\linewidth]{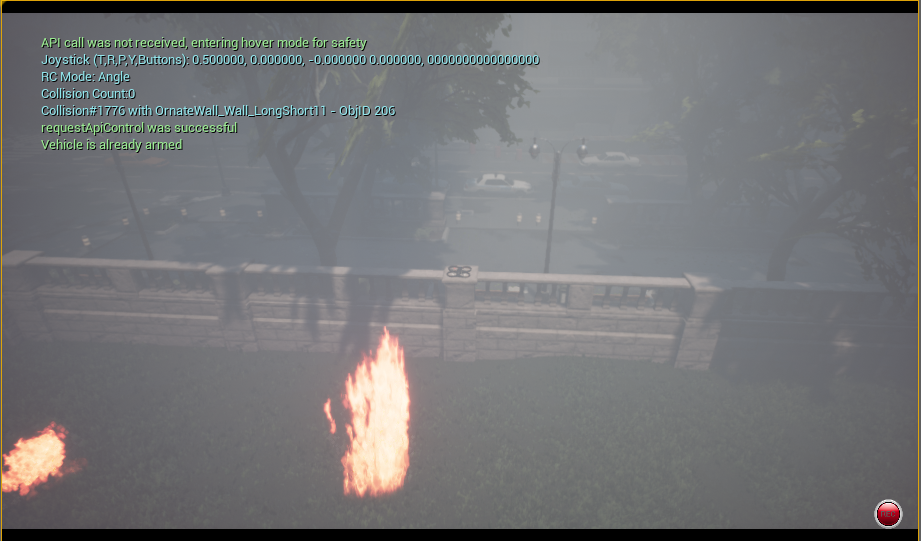}\\
          
               \includegraphics[width=0.51\linewidth]{Images/pleaf5.png}
          \includegraphics[width=0.475\linewidth]{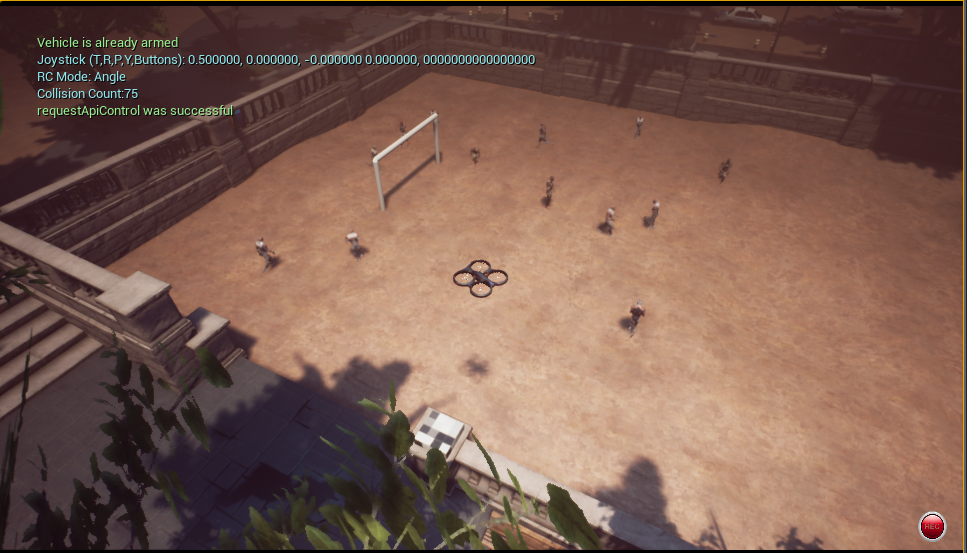}
}}
    
      \caption{Images sampled from the modified \textit{Downtown} simulated environments for fine-tuning and meta-learning}
      \label{downtownsamples2}
   \end{figure}

A key challenge for supervised navigation algorithms is the generalizability to handle changing, dynamic environments. 
Given that it is impossible to generate training data that cover all possible situations the UAV can encounter, we aim to build a meta-learning/adaptive algorithm that allows the model to predict the pose of the vehicle in unseen conditions.  
Fig.~\ref{downtownsamples2} illustrates samples of perturbed \textit{Downtown} environment.
Each environment consists of the same central structure, i.e., a grass field surrounded by pillars, with different weather conditions, materials, and lights. 

\paragraph{Loss function for ego-motion consistency}
From the results, we observe that in some cases, the estimated pose is not consistent with the previous estimate. 
One possible solution is to use prior information about system dynamics at the previous time step to constrain the prediction of the next estimated pose.
Hence, future work may be focused on deriving a loss function that incorporates previous motion information to guarantee consistency between continuous streams of data.

\section{Conclusion}

We propose a new end-to-end learning method for pose estimation, and we assest the performances of our model with state-of-the-art methods for visual-inertial estimation on the EuRoC MAV benchmark dataset. The results show that our deep learning approach outperforms the baseline compared to its feature-based counterparts.
We finally integrate our estimator in the Airsim closed-loop control system, and we demonstrate in a simulation that our data-driven policy can navigate and land a UAV autonomously on its target with less than 10cm of error. 

\bibliography{bibliography} 
\bibliographystyle{IEEEtran}

\end{document}